\documentclass{bmvc2k}

\usepackage{enumitem}
\usepackage{booktabs}
\usepackage{multirow}

\usepackage{comment}
\usepackage{wrapfig}

\usepackage{floatrow}
\newfloatcommand{capbtabbox}{table}[][\FBwidth]

\title{Black Magic in Deep Learning: How Human Skill Impacts  Network Training}

\addauthor{Kanav Anand}{anandkanav92@gmail.com}{1}
\addauthor{Ziqi Wang}{z.wang-8@tudelft.nl}{1}
\addauthor{Marco Loog}{M.Loog@tudelft.nl}{12}
\addauthor{Jan van Gemert}{j.c.vangemert@tudelft.nl}{1}

\addinstitution{
Delft University of Technology,\\
Delft, The Netherlands
}

\addinstitution{
 University of Copenhagen\\
 Copenhagen, Denmark
}

\runninghead{Anand, Wang, Loog, van Gemert}{Black Magic in Deep Learning}

\def\etal{\emph{et al}\bmvaOneDot}
\def\etc{\emph{etc}\bmvaOneDot}
\def\ie{\emph{i.e}\bmvaOneDot}

\begin{document}

\maketitle

\begin{abstract}

How does a user's prior experience with deep learning impact accuracy?  We present an initial study based on 31 participants with different levels of experience. Their task is to perform hyperparameter optimization for a given deep learning architecture. The results show a strong positive correlation between the participant's experience and the final performance. They additionally indicate that an experienced participant finds better solutions using fewer resources on average. The data suggests furthermore that participants with no prior experience follow random strategies in their pursuit of optimal hyperparameters. Our study investigates the subjective human factor in comparisons of state of the art results and scientific reproducibility in deep learning. 
\end{abstract}

\section{Introduction}
\label{sec:intro}

The popularity of deep learning in various fields such as image recognition~\cite{image_1,image_2}, speech~\cite{speech_1,speech_2}, bioinformatics~\cite{bioinformatics,drug},  question answering~\cite{qa} \etc. stems from the seemingly favorable trade-off between the recognition accuracy and their optimization burden. LeCun~\etal~\cite{deeplearning} attribute their success to  feature representation learning as opposed to using hand-engineered features. While deep networks learn features, the hand engineering has shifted to the design and optimization of the networks themselves. In this paper we investigate the influence of human skill in the hand engineering of deep neural network training.

Arguably, one reason for why neural networks were less popular in the past is that compared to `shallow' learners such as for example LDA~\cite{fisher1936use}, SVM~\cite{cortes1995support}, $k$NN~\cite{cover1967nearest}, Naive-Bayes~\cite{rish2001empirical}, \etc, deep networks have many more hyperparameters~\cite{zhou2018exploring} such as the number of layers, number of neurons per layer, the optimizer, optimizer properties, number of epochs, batch size, type of initialization, learning rate, learning rate scheduler, \etc. A hyperparameter  has to be set before training the deep network and setting these parameters can be difficult~\cite{smith2018disciplined}, yet, the excellent results of deep networks~\cite{deeplearning} as revealed by huge datasets~\cite{imagenet} with fast compute~\cite{image_1} offer a compelling reason to use deep learning approaches in practice, despite the difficulty of setting many of those parameters. 

Hyperparameters are essential to good performance as many learning algorithms are critically sensitive to hyperparameter settings~\cite{fanova, koutsoukas2017deep, reimers2017optimal}. The same learning algorithm will have different optimal hyperparameter configurations for different tasks~\cite{diff_datasets} and optimal configurations for one dataset do not necessarily translate to others~\cite{across_datasets}. The existing state of the art can be improved by  reproducing the work with a better analysis of hyperparameter sensitivity~\cite{prac-bayesian}, and several supposedly novel models in NLP~\cite{nlp-reproduce} and in GANs~\cite{gans} were found to perform similarly to existing models, once hyperparameters were sufficiently tuned. These results show that hyperparameters are essential for reproducing existing work, evaluating model sensitivity, and making comparisons between models.

Finding the best hyperparameters is something that can be done
automatically by autoML~\cite{bergstra2013making, domhan2015speeding, kerschke2019automated, kotthoff2017autoWEKA} or Neural Architecture Search~\cite{elsken2018neural, liu2018darts, pham2018efficient, zoph2016neural}. Yet, in practice, such methods are not widely used by deep learning researchers. One reason could be that automatic methods are still under active research and not yet ready for consumption. Another reason could be that good tuning adds a significant computational burden~\cite{gans,nlp-reproduce}.  Besides, automated tuning comes with its own set of hyperparameters and, in part, shifts the hyperparameter problem.  Thus, in current practice, the hyperparameters are usually set by the human designer of the deep learning models. In fact, it is widely believed that hyperparameter optimization is a task reserved for experts~\cite{taking-humanout-of-loop, smith2018disciplined}, as the final performance of a deep learning model is \emph{assumed} to be highly correlated with background knowledge of the person tuning the hyperparameters. The validation of this claim is one of the main goals of our research. The extraordinary skill of a human expert to tune hyperparameters is what we here informally refer to as ``black magic'' in deep learning.

\subsection{Contributions}

Broadly speaking, we investigate how human skill impacts network training. More specifically, we offer the following contributions. 1. We conduct a user study where participants with a variety of  experience in deep learning perform hyperparameter optimization in a controlled setting.\footnote{The research carried out has been approved by TU Delft’s Human Research Ethics Committee:\\ \url{https://www.tudelft.nl/en/about-tu-delft/strategy/integrity-policy/human-research-ethics/}.}
2. We investigate how deep learning experience correlates with model accuracy and tuning efficiency. 3. We investigate human hyperparameter search strategies. 4. We provide recommendations for reproducibility, sensitivity analysis, and model comparisons.






\section{Experimental Setup}

Our experiment is designed to measure and analyze human skill in hyperparameter optimization. All other variations have identical settings. Each participant has the exact same task, model, time limitation, GPU, and even the same random seed. Our participants tune  hyperparameters of a deep learning architecture on a given task in a user-interface mimicking a realistic setting while allowing us to record measurements. 

\subsection{Deep Learning Setup}

The deep learning experimental setup includes: the task, the model and the selection of hyperparameters.

\paragraph{Deep learning task.} The choice for the task is determined by the size, difficulty, and realism of the considered dataset. Large datasets take long to train, which limits the number of hyperparameters we can measure.  Also, if the dataset is not challenging, it would be relatively easy to achieve a good final performance which limits the variance in the final performance of the model. Taking size and difficulty into account, while staying close to a somewhat realistic setting, we decided on an image classification task on a subset of ImageNet~\cite{imagenet} which is called \textit{Imagenette}~\cite{imagenette}. To prevent using dataset specific knowledge we did not reveal the dataset name to participants.  We only revealed the image classification task and we shared the dataset statistics: 10 classes, 13,000 training images, 500 validation images, and 500 test images

\paragraph{Deep learning model.} The model should be well-suited for image classification, have variation in hyperparameter settings, and be somewhat realistic.  In addition, it should be relatively fast to train so that a participant can run a reasonable amount of experiments in a reasonable amount of time.  We selected \textit{Squeezenet}~\cite{squeezenet} as it is efficient to train and achieves a reasonable  accuracy compared to more complex networks. To prevent exploiting model-specific knowledge, we did not share the network design with the participants.

\begin{figure}
\begin{floatrow}
\ffigbox{%
  \includegraphics[width=\linewidth]{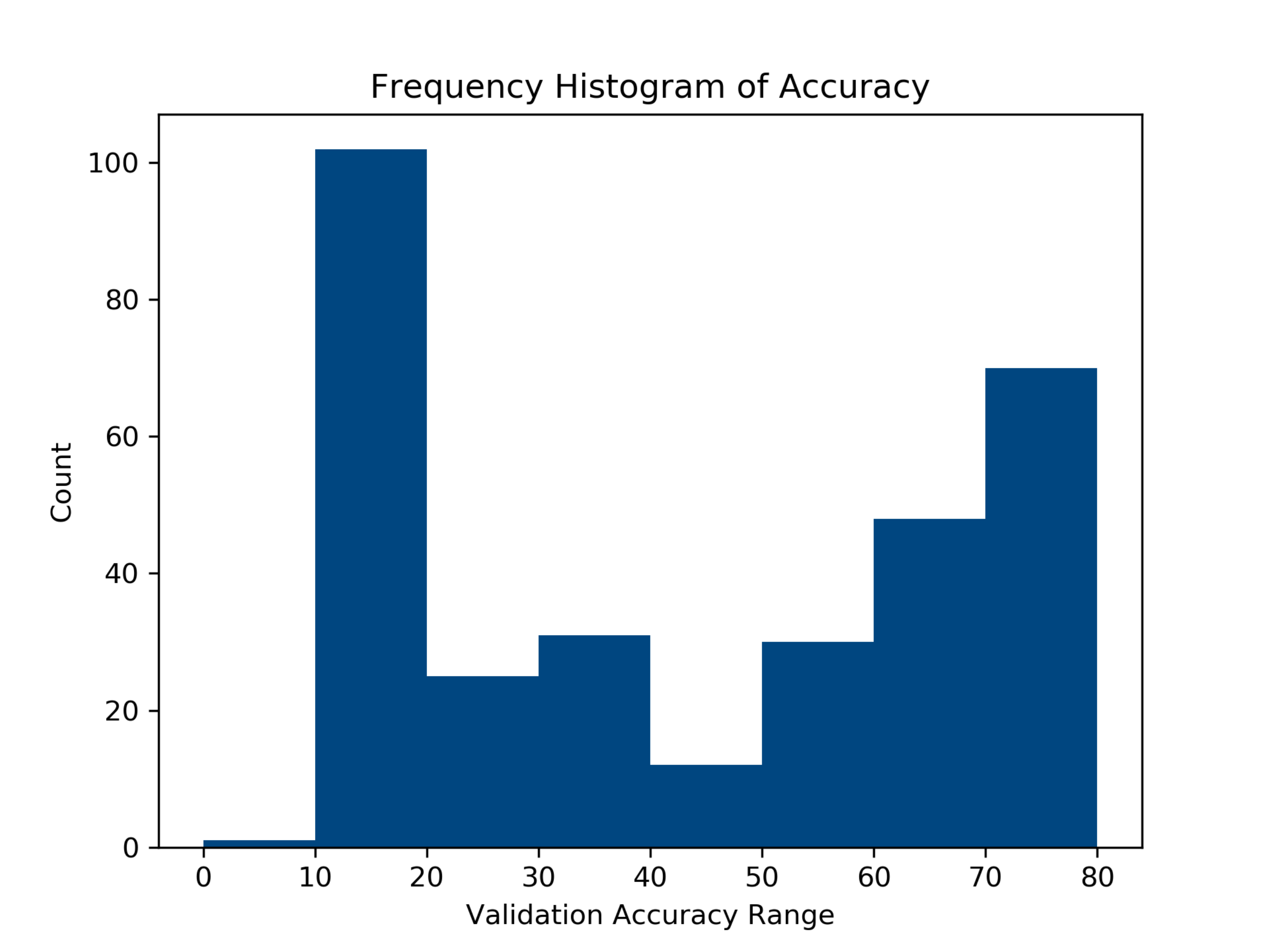}%
}{%
  \caption{Accuracy histogram over 320 random hyperparameter settings. Their settings matter. }%
  \label{fig:hp_hist}
}
\capbtabbox{%
\resizebox{0.9\linewidth}{!}{%
\begin{tabular}{@{}lll@{}}
\toprule
    \textbf{Type} & \textbf{Hyperparameter} & \textbf{Default value}\\ \hline
   \multirow{4}{*}{Mandatory} & Epochs & - \\ 
    &   Batch size & - \\
    &   Loss function & - \\
    &   Optimizer & - \\
    \cmidrule(r){2-3}
    \multirow{11}{*}{Optional} & Learning rate & 0.001 \\
    &   Weight decay & 0 \\
    &   Momentum & 0 \\
    &   Rho & 0.9 \\
    &   Lambda & 0.75 \\
    &   Alpha & 0.99 \\
    &   Epsilon & 0.00001 \\
    &   Learning rate decay & 0 \\
    &   Initial accumulator value & 0 \\
    &   Beta1 & 0.9 \\
    &   Beta2 & 0.999 \\
    \bottomrule    
\end{tabular}
}
}{%
  \caption{The hyperparameters available to participants in our study.}%
  \label{tab:hp_list}%
}
\end{floatrow}
\end{figure}

\paragraph{Hyperparameters.} We give participants access to 15  common hyperparameters. Four parameters are mandatory: number of epochs, batch size, loss function, and optimizer. We preset the other 11 optional hyperparameters with their commonly used default values. In Table~\ref{tab:hp_list}, we show the list of hyperparameters. Please refer to the supplementary material for their full description. Note that none of the hyperparameters under participants control influenced the random seed, as we keep any randomness such as weight initialization and sample shuffling exactly the same for each participant. For 320 random hyperparameter settings, the average random accuracy is $41.8 \pm 24.3$, where Figure~\ref{fig:hp_hist} demonstrate that hyperparameters are responsible for ample accuracy variance for this task. Without such variance there may be little differences in human accuracy which would make it difficult to analyse skill.


\subsection{Participants' Experimental Setup}

For managing participants we need: a user-interface, a detailed task description, and define what to measure.

\begin{figure}
\centering
\includegraphics[width=0.8\paperwidth]{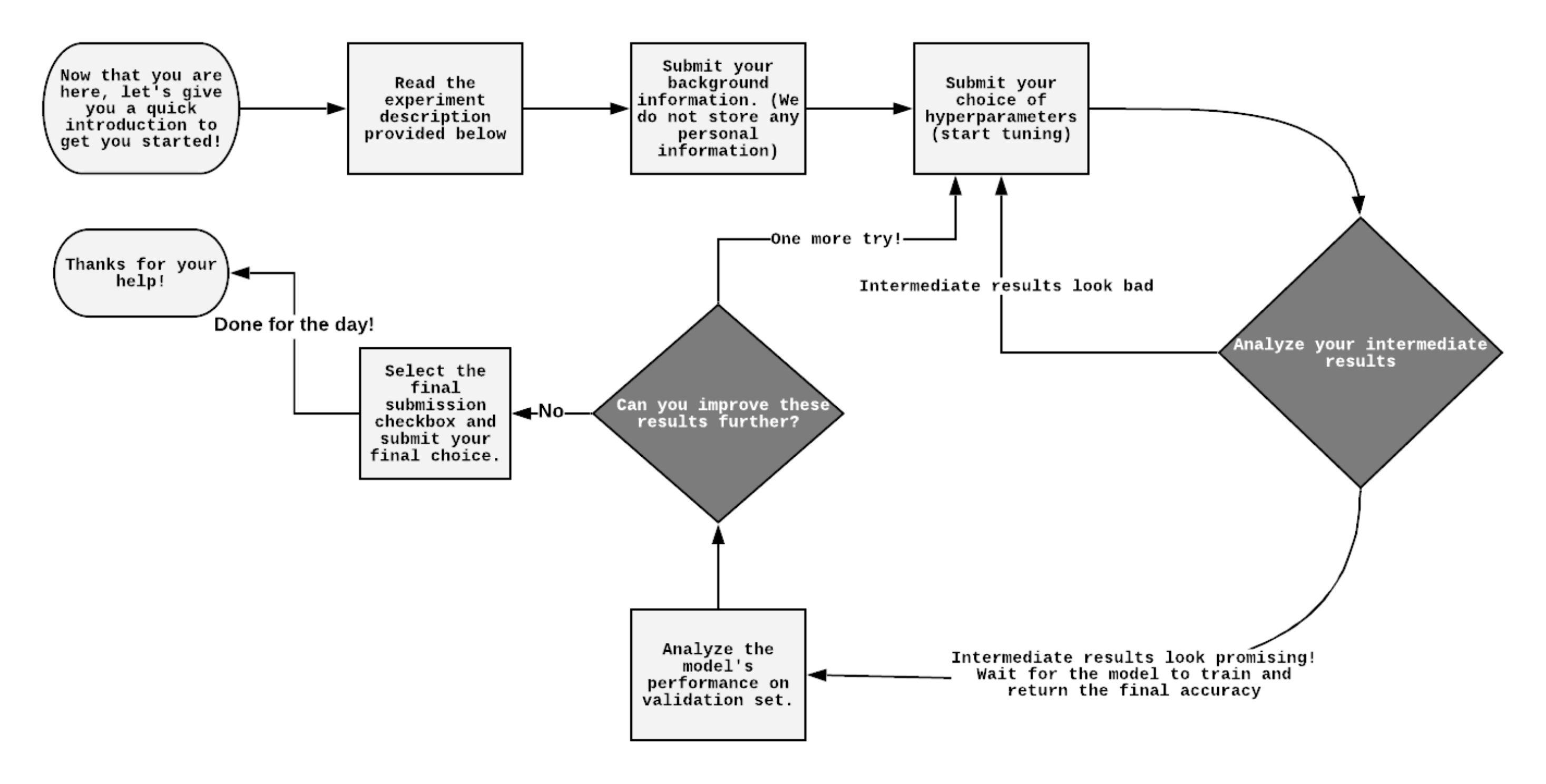}
\caption{The flow diagram of the user study. The participant starts by entering their information. Next, submit the values for hyperparameters and evaluate intermediate training results. If the training is finished, the participant can decide whether to submit a new configuration for hyperparameter or end the experiment. It can be repeated until the time limit of 120 minutes is reached.}
\label{fig:flow}
\end{figure}

\paragraph{User-interface.} We simulate a realistic hyperparameter optimization setting, while providing a controlled environment. We designed a web interface to let participants submit their choice of hyperparameters, view their submission history with validation accuracy and view the intermediate training results with an option for early stopping. Few preliminary tries were done (by the participants not included in result dataset) to test and verify the design and hyperparameter optimization process. By using a web server we collect all the data for analysis. We make all data and source code available\footnote{\url{https://github.com/anandkanav92/htune}}.

\paragraph{Participant's task.} The task given to participants is to find the optimal set of hyperparameters, \ie{}, those maximizing classification accuracy on the test set. After submitting a choice of hyperparameters, the deep learning model is trained in the background using these parameters. While the model is training, the participant can view the intermediate batch loss and epoch loss in real time. The participant has an option to cancel training if the intermediate results do not look promising. As there is an upper limit of 120 minutes to how much time a participant can use on the optimization of the model, early stopping enables them to try more hyperparameter configurations. After training the model is finished, the accuracy on a validation set is provided to the participant. Participants are encouraged to add optional comments to each choice of hyperparameters. The experiment ends when the participant decides that the model has reached its maximum accuracy or if the time limit of the experiment is reached (120 minutes). The flow diagram of the user study is depicted in Figure~\ref{fig:flow}.

\paragraph{Measurements per participant.} As an indication for the degree of expertise a participant has, we record the number of months of deep learning experience. During deep model training, we record all the hyperparameter combinations tried by the participant, together with the corresponding accuracy on the validation set, for as many epochs as the participant chooses to train. The experiment ends by a participant submitting their final choice of hyperparameters. This optimal hyperparameter configuration is then trained ten times on the combined training and the validation set after which the accuracy on the independent test set is recorded. Each of the 10 repeats have a different random seed, while the seeds are the same for each participant.

\subsection{Selection of Participants}
The participants were selected based on their educational background and their prior experience in training deep learning models. The participants with no prior experience comprised of people recruited from different specialisations using poster ads and email groups. Experienced candidates were invited through our deep learning course provided to master students and researchers. 
 


\section{Results}\label{sect:res}
\begin{wrapfigure}{R}{0.6\textwidth}
\centering
\includegraphics[width=0.9\textwidth]{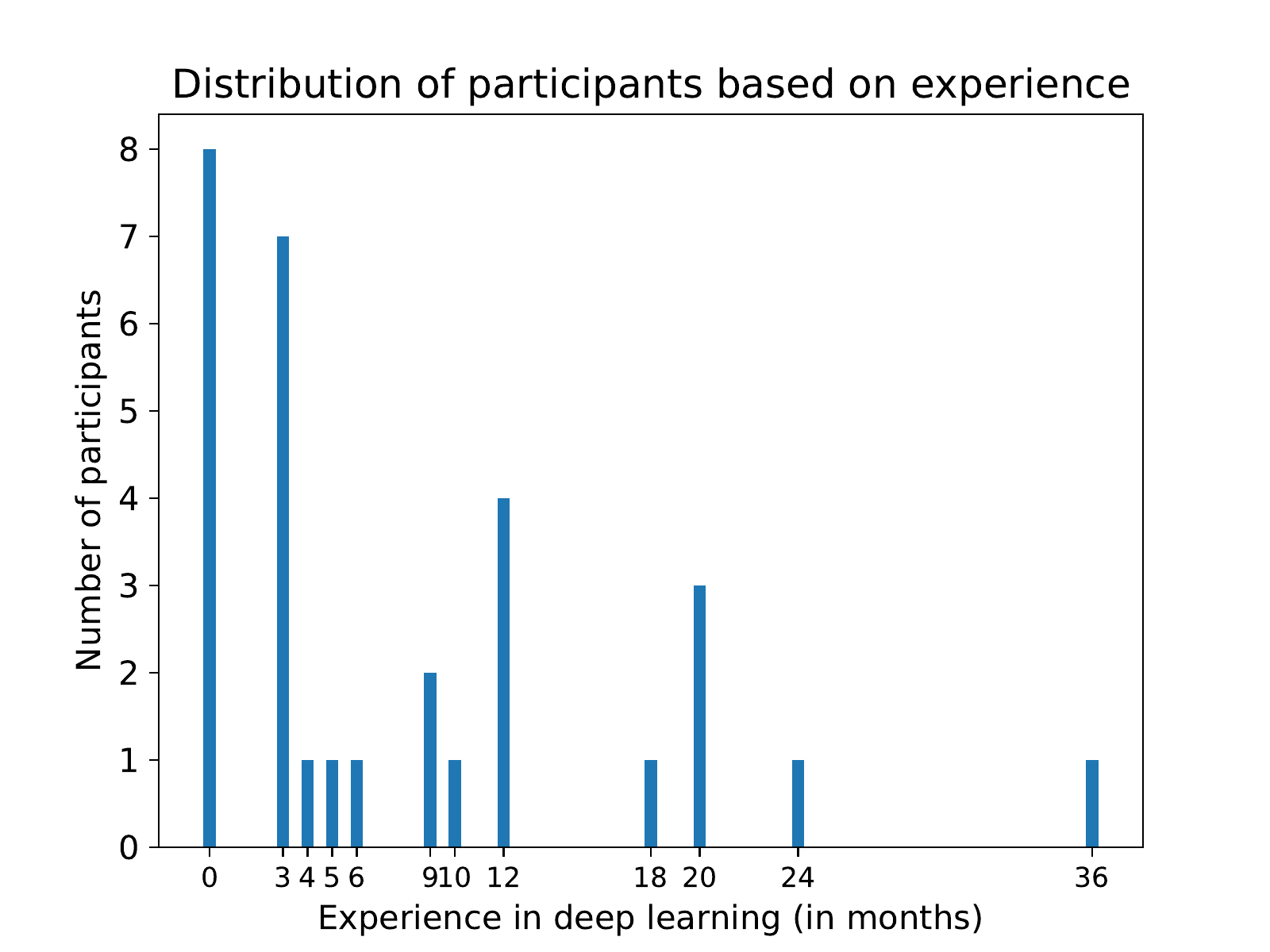}
\caption{A broad range of  deep learning experience in the 31 participants of our study.}
\label{fig:hist_part}
\end{wrapfigure}

We collected 463 different hyperparameter combinations from 31 participants. The prior deep learning experience for these participants is well distributed as shown in Figure~\ref{fig:hist_part}. For the final selected hyperparameters the average classification accuracy is $55.9 \pm 26.3$.

For ease of analysis we  divide participants into groups based on experience. The \textit{Novice} group contains 8 participants with no experience in deep learning, the 12 participants in the \textit{medium} group have less than nine months of experience and the 11 participants in the \textit{expert} group has more than nine months experience. 




\subsection{Relation between Experience and Accuracy}
\begin{figure}
\begin{floatrow}
\ffigbox{%
  \includegraphics[width=\linewidth]{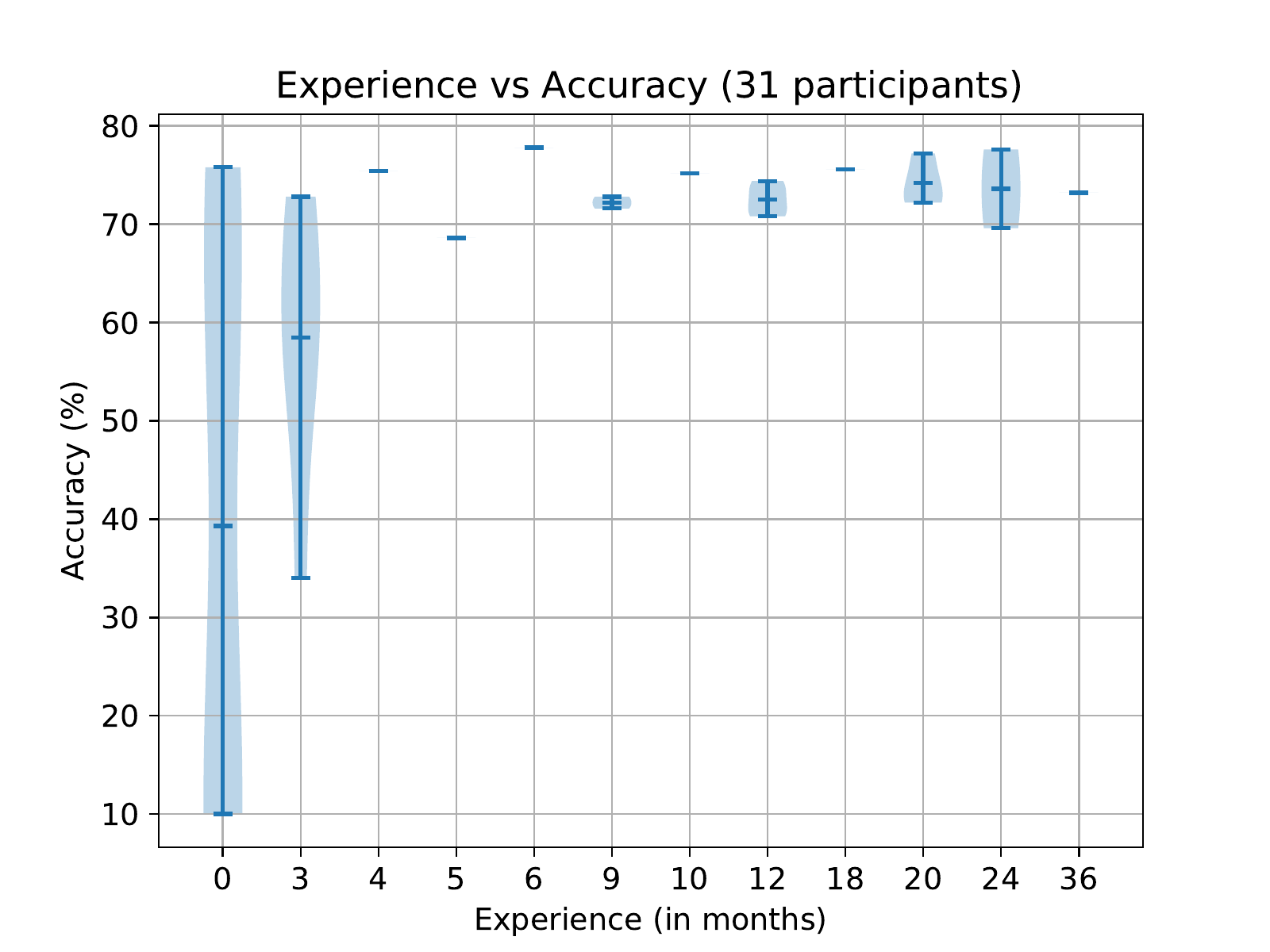}%
}{%
  \caption{Final accuracy distribution over all participants.  }%
  \label{fig:group_acc}
}
\ffigbox{%
  \includegraphics[width=\linewidth]{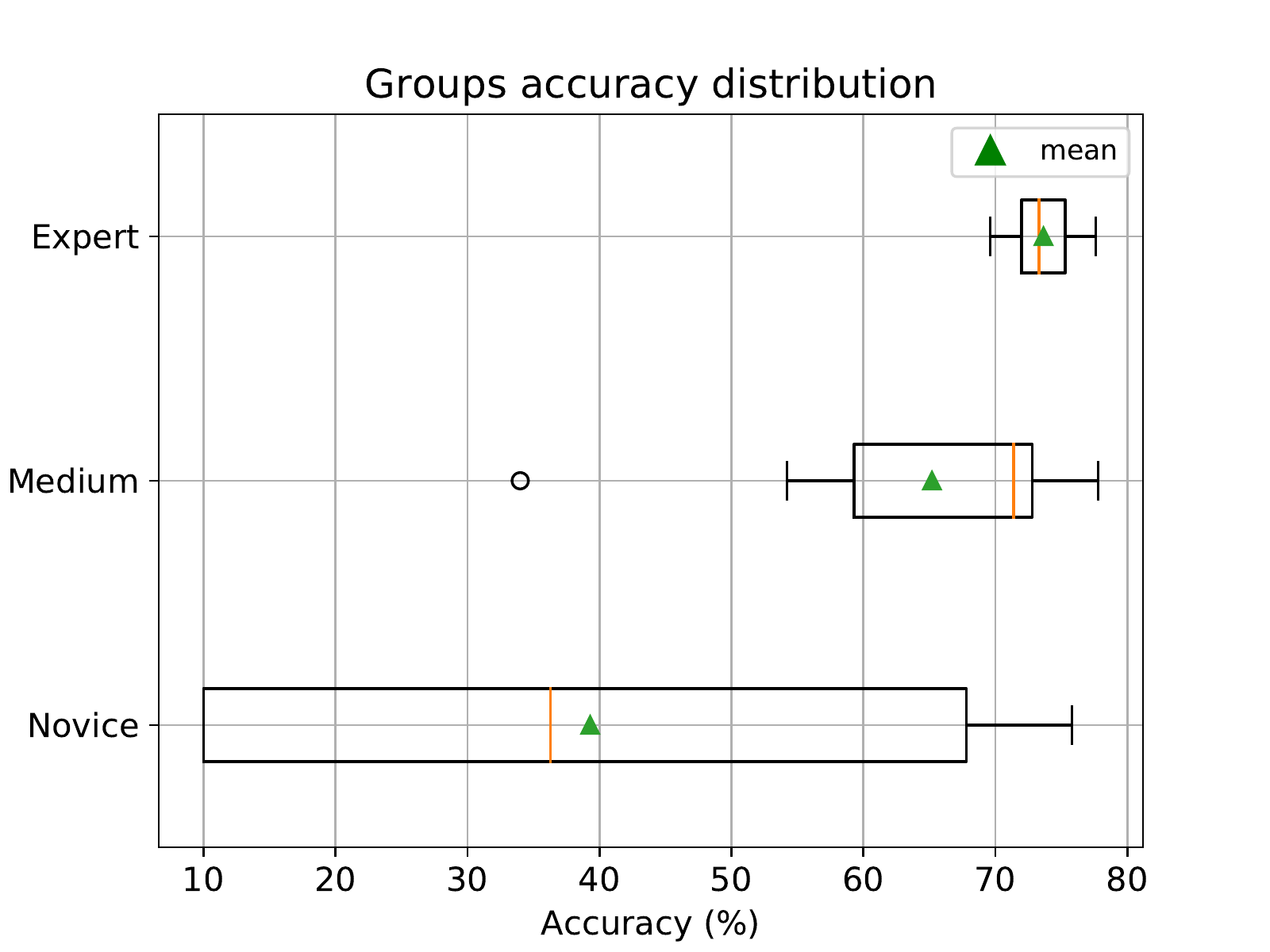}%
}{%
  \caption{Final accuracy per group boxplot. }%
  \label{fig:box_plot_acc}
}
\end{floatrow}
\end{figure}

\begin{figure}
\begin{floatrow}
\capbtabbox{%
\resizebox{0.9\linewidth}{!}{%
\begin{tabular}{ccc}
\toprule
\multicolumn{3}{c}{Levene's statistical test } \\
 \textbf{Groups}&
 \textbf{Test Statistic}
 & \textbf{p-value}  \\
\midrule
    Novice vs Medium &  8.40 & 0.01\\ 
     Novice vs Expert & 14.338 & 0.001\\ 
     Medium vs Expert & 5.52 & 0.029\\ \bottomrule \\ \\ \\
\end{tabular} 
}
}{%
  \caption{ All groups significantly differ from each other ($p<0.05$);  \textit{medium} and \textit{expert} the least and \textit{Novice} and \textit{expert} the most. }%
  \label{tab:levene}%
}
\ffigbox{%
  \includegraphics[width=\linewidth]{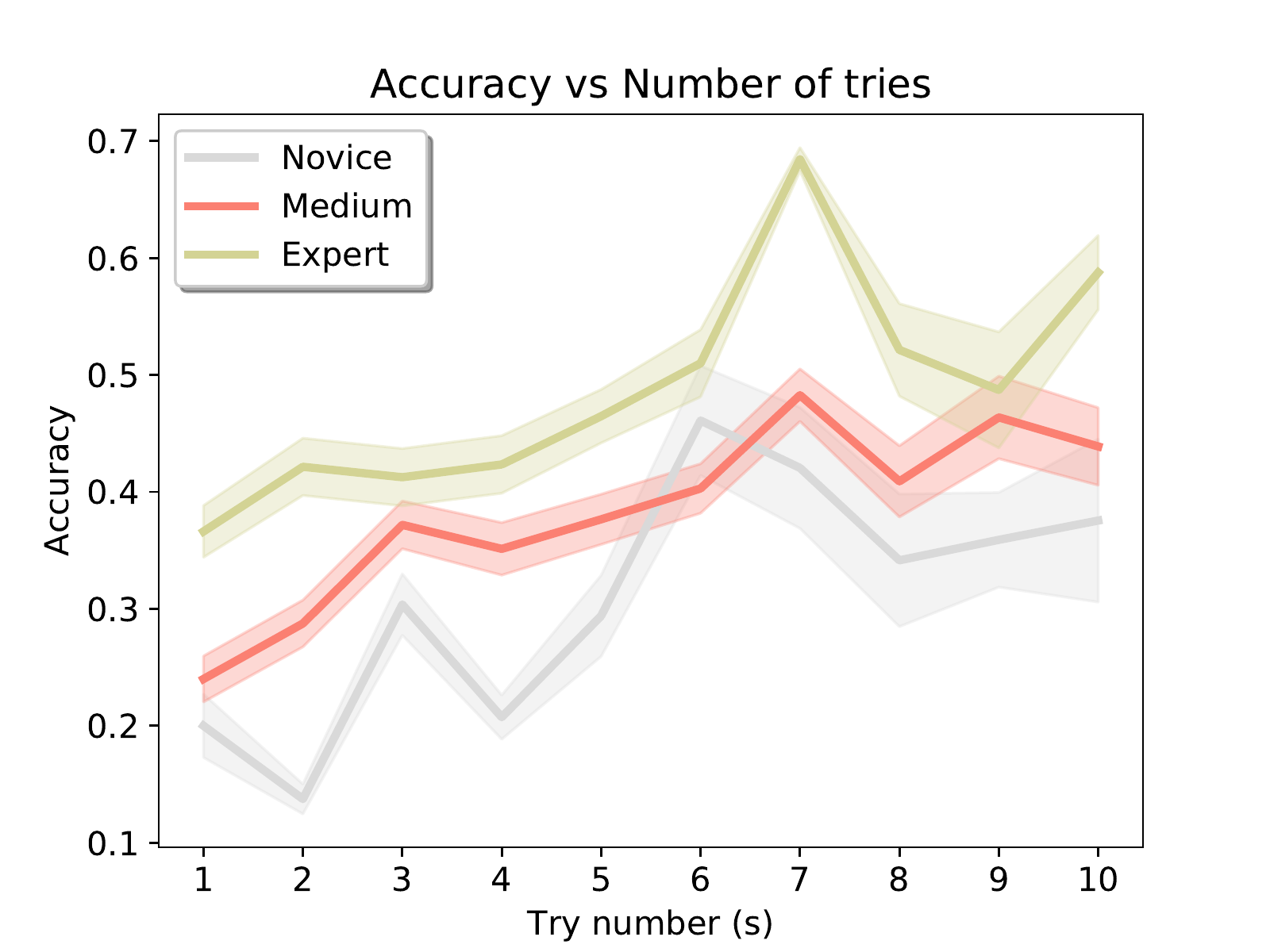}%
}{%
  \caption{Experts need fewer tries to get better accuracy. The shaded region is standard error. }%
  \label{fig:trace}
}
\end{floatrow}
\end{figure}

Figure~\ref{fig:group_acc} depicts the relationship between final accuracy and deep learning experience per participant. As the experience increases, the final accuracy tends to increase, which is supported by the strong positive Spearman~\cite{spearman} rank order correlation coefficient of 0.60 with a $p$-value smaller than 0.001. 
Additionally, we compared the variance of the accuracy distributions  of \textit{Novice},   \textit{medium},  \textit{expert} groups using Levene's statistical test~\cite{levene}. We use the Levene test because experience and accuracy are not normally distributed. The test values presented in Table~\ref{tab:levene} show all groups significantly differ from each other ($p<0.05$), where the difference is smallest between \textit{medium} and \textit{expert} and the largest  between \textit{Novice} and \textit{expert}, which is in line with the accuracy statistics per group shown in Figure~\ref{fig:box_plot_acc}.

\begin{figure}
\begin{tabular}{@{}c@{}c@{}}
\includegraphics[width=0.5\columnwidth,keepaspectratio]{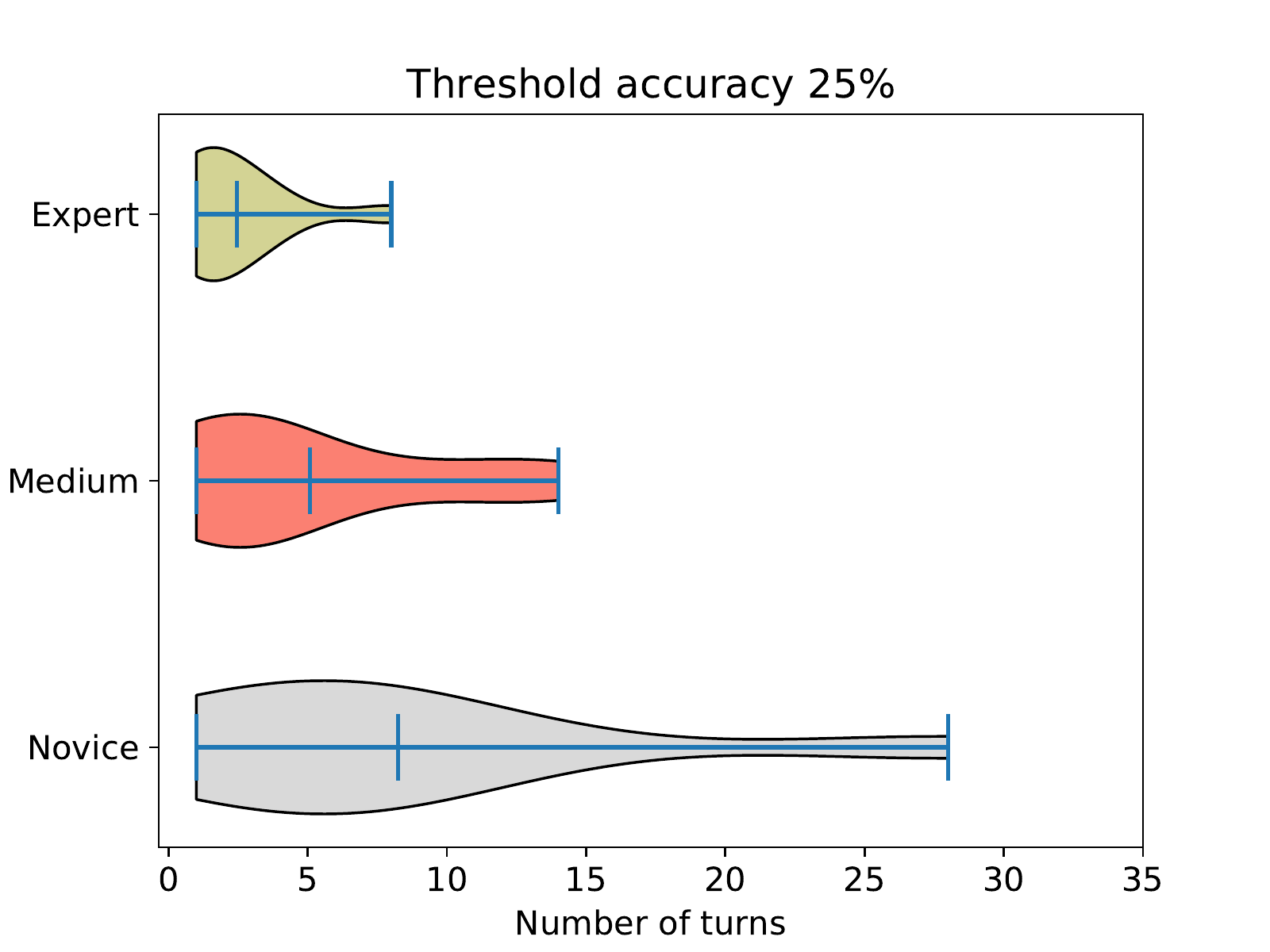} &
\includegraphics[width=0.5\columnwidth,keepaspectratio]{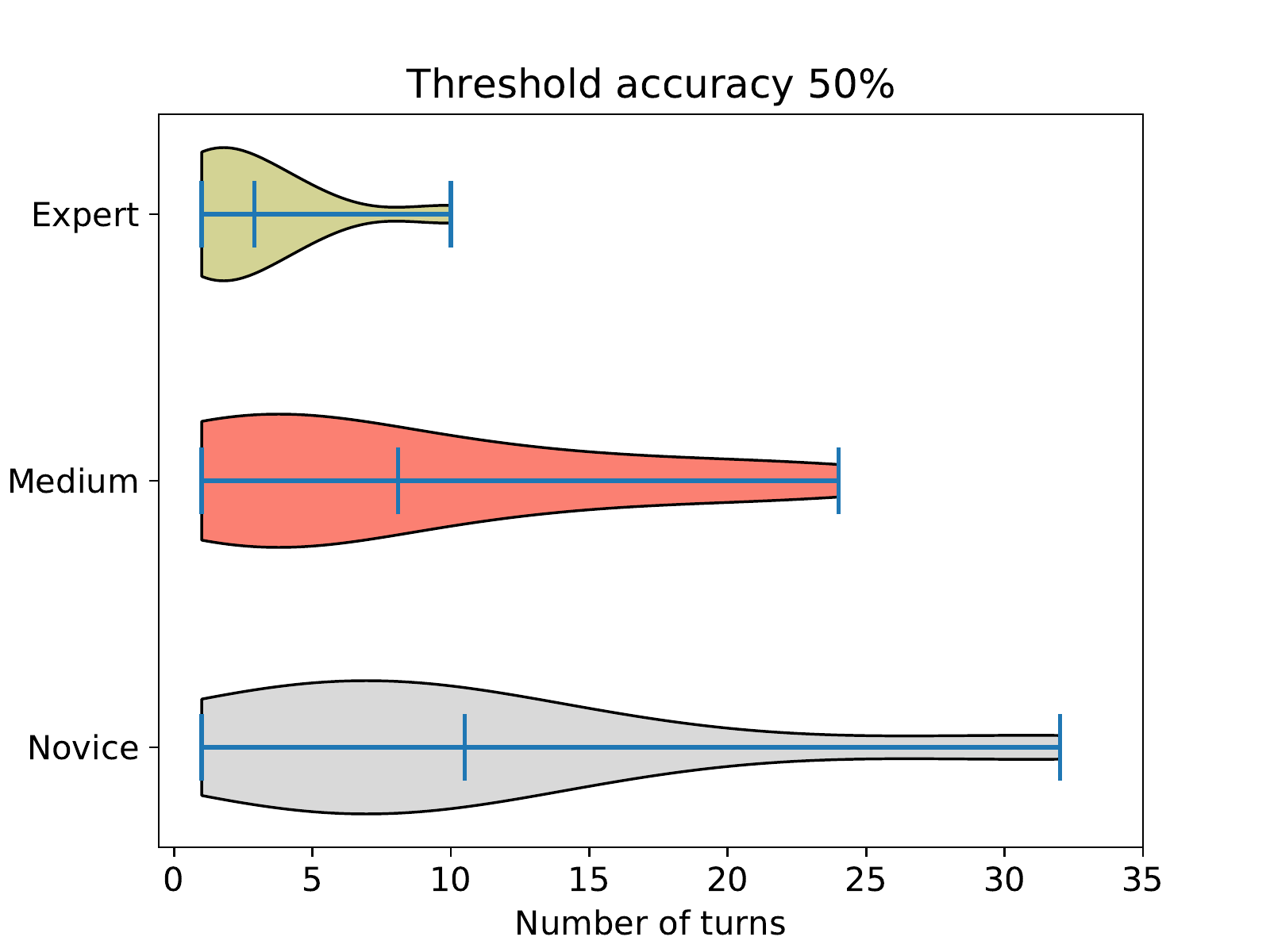} 
\end{tabular}
\caption{Number of hyperparameter configurations required to achieve a threshold accuracy of 25\%, 50\% for different experience groups. The violin plot shown above depicts the probability density distribution of the number of turns taken by the participants in each group. The mean value of each group is marked for reference. More experienced participants reach the threshold faster. }
\label{fig:quick}
\end{figure}

We further analyze the effect of deep learning experience on the training process. In Figure~\ref{fig:quick}, we show how many tries are used to reach a certain threshold accuracy for the \textit{novice},  \textit{medium}, \textit{expert} groups for final accuracy thresholds. Experts reach the threshold quicker. Furthermore, we show the average accuracy of each group after a number of tries in Figure \ref{fig:trace}. We can conclude that more experienced participants not only achieve a better accuracy, they also arrive at that better score more efficiently.



\subsection{Difference in Strategies}

We investigate why more experienced users achieve a higher accuracy in fewer iterations. 

\begin{figure}
\begin{floatrow}
\ffigbox{%
  \includegraphics[width=\linewidth]{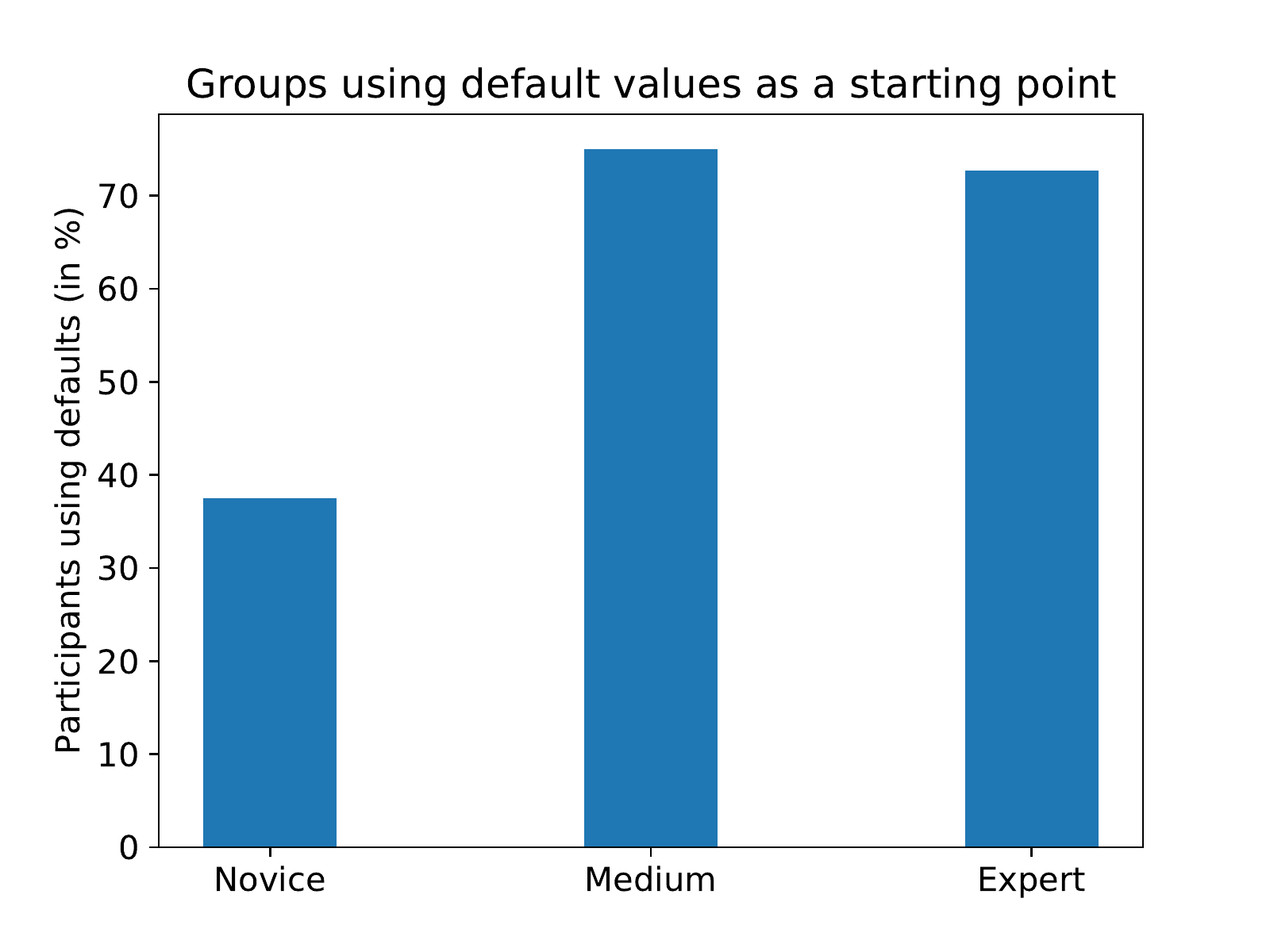}%
}{%
  \caption{Participants submitting their initial hyperparameter configuration using all default values.}%
  \label{fig:default}
}
\capbtabbox{%
\resizebox{0.9\linewidth}{!}{%
\begin{tabular}{rp{5cm}}
\toprule
 \textbf{ID} & \textbf{Comment}  \\
\midrule
    1 & It is just a guess. \\ 
    2 & It is a suggested default value. \\ 
    3 & It is the value that has worked well for me in the past.\\ 
    4 & It is the value I learnt from previous submissions.\\ 
    5 & Other \\ \bottomrule \\ 
\end{tabular}
}
}{%
  \caption{Predefined comments used in user study.}%
  \label{tab:comments}%
}
\end{floatrow}
\end{figure}

\paragraph{Use of suggested default values.} We offer mandatory and optional hyperparameters, as shown in Table \ref{tab:hp_list}, where the optional hyperparameters are preset to their default values. Figure \ref{fig:default} shows the number of participants in each group using these default values as the starting point. A large majority in the \textit{medium} or \textit{expert} groups begin with all optional hyperparameter values set to their suggested default values and subsequently build on them. In contrast, \textit{novice} users directly explore the optional values. Using defaults for optional parameters does not necessarily lead to an optimal hyperparameter configuration, however, all participants who started with defaults achieved a final performance greater than 50\%.


\begin{figure}
    \centering
    \includegraphics[width=\textwidth]{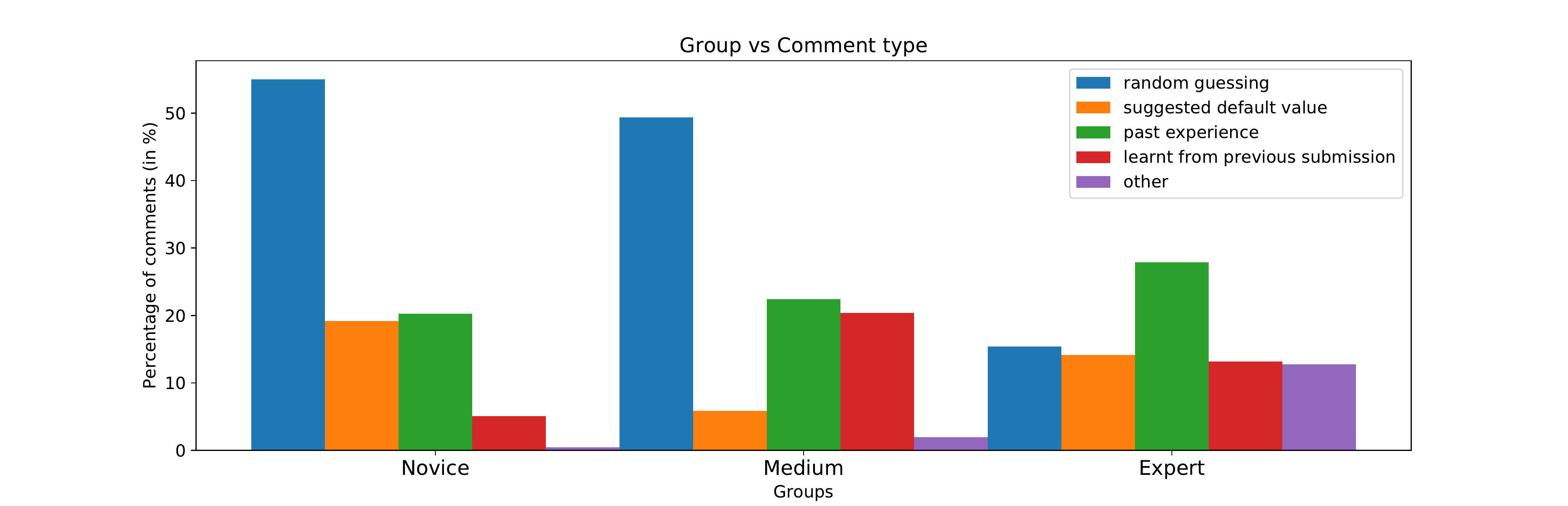}
    \caption{The distribution of comments for the groups of \textit{novice}, \textit{medium},  \textit{expert} participants. Inexperienced users rely more on random guessing (blue). }
    \label{fig:groupComments}
\end{figure}

\paragraph{Analysis of comments.} Participants were encouraged to leave comments explaining the reasoning behind choosing a specific value of a hyperparameter. In a bid to gather maximum comments, we let users choose from predefined comments shown in Table \ref{tab:comments}. Figure \ref{fig:groupComments} shows the distribution of comments for each group of \textit{novice},  \textit{medium}, or \textit{expert} participants. We noticed that there was confusion between `past experience' and `learned from previous submission' as 22\% of hyper parameter values used by \textit{novice} participants were based on their prior experience in deep learning.  As this confusion may also effect other groups, we refrain from drawing hard conclusions based on the observed increase in the use of the comment `past experience' for more experienced participants. For \textit{novice} participants, the majority is based on random guessing. Random guessing was found to be strongly negatively correlated with the increasing experience. We used Spearman rank-order correlation, and the value was found to be $-0.58$ with a $p$-value smaller than 0.001. As the amount of experience increases, the results show a decrease in random guessing.



\section{Discussion and Conclusion}

We identify main limitations to this study, draw conclusions, and make recommendations.

\subsection{Main Limitations}

\paragraph{Limited data.} We have a fairly restricted number of 31 participants. Collecting more data and inviting more participants in the user study will make the result and conclusions more robust to potential outliers. In addition, it can of course provide better insight into the process of hyperparameter optimization, generalize our findings over a broader audience, and give us the possibility to test more refined hypotheses.

\paragraph{Stratified experience groups.} Currently, the three participant groups that we used in our analysis, \ie{}, novice, medium, and expert, were identified based on the amount of experience, as measured months, they had.  It may be of interest, of course, to consider information different from experience to stratify participants in different groups.  Maybe the amount of programming experience or the amount of machine learning experience correlates better with performance achievements. What should maybe also be considered, however, is the way to measure something like experience.  Rather than using a measure like `months of experience,' one can also resort, for instance, to often used self-evaluations, in which every participant decided for themselves which level they have.  In more extensive experiments, it would definitely be of interest to collect such additional meta-data.


\paragraph{Only one deep learning setting} This study focuses only on an image recognition task with a single model and a single dataset in a limited time. Thus, it can be argued that the findings of this study could not be generalized to other deep learning settings. This work is the first study explicitly analyzing human skill in hyperparameter tuning; it is interesting to extend this study further by including multiple tasks, models and datasets.

\subsection{Conclusions}

\paragraph{Human skill impacts accuracy.} Through this user study, we found for people with similar levels of experience tuning the exact same deep learning model, the model performs differently. Every source of variation was eliminated by fixing the task, the dataset, the deep learning model, and the execution environment (random seed, GPUs used for execution) except the choice of hyperparameters. Figure \ref{fig:box_plot_acc} shows the variance in the final performance of the model. This suggests that final performance of the model is dependent on the human tuning it. Even for experts the difference can be an accuracy difference of 5\%. 

\paragraph{More experience correlates with optimization skill.}
We show a strong positive correlation between experience and final performance of the model. Moreover, the data suggests that more experienced participants achieve better accuracy more efficiently, while inexperienced participants follow a random search strategy, where they often start by tuning optional hyperparameters which may be best left at their defaults initially.

\subsection{Recommendations}

Concluding our work, we would like to take the liberty to propose some recommendations regarding experiments and their outcome. We base these recommendations on our observed results that even expert accuracy can differ as much as 5\% due to hyperparameter tuning. Thus, hyperparameters are essential for reproducing the accuracy of existing work, for making comparisons to baselines, and for making claims based on such comparisons.

\begin{itemize}

\item Reproducability: Please share the final hyperparameter settings. 


\item Comparisons to baselines: Please optimize and report the hyperparameter settings for the baseline with equal effort as the proposed model.

\item Claims of (the by now proverbial) superior performance: It is difficult to say if the purported superior performance is due to a massive supercomputer trying all settings~\cite{gans,nlp-reproduce}, due to a skilled human as we show here, or due to qualities of the proposed model. Bold numbers correlate with black magic and we recommend to make bold numbers less important for assessing the contribution of a research paper. 

\item To the deep learning community: Make reviewers pay more attention to reproducability, baseline comparisons, and put less emphasis on superior performance. There is no need to burn  wielders of black magic at the stake, but herald the enlightenment by openness and clarity in hyperparameter tuning.

\end{itemize}

\section*{Acknowledgement}
This work is part of the research programme C2D–Horizontal Data Science for Evolving Content with project name DACCOMPLI and project number 628.011.002, which is (partly) financed by the Netherlands Organisation for Scientific Research (NWO).

\bibliography{ms}

\begin{thebibliography}{36}
\providecommand{\natexlab}[1]{#1}
\providecommand{\url}[1]{\texttt{#1}}
\expandafter\ifx\csname urlstyle\endcsname\relax
  \providecommand{\doi}[1]{doi: #1}\else
  \providecommand{\doi}{doi: \begingroup \urlstyle{rm}\Url}\fi

\bibitem[ima()]{imagenette}
Imagenette.
\newblock \url{https://github.com/fastai/imagenette}.

\bibitem[Bergstra et~al.(2013)Bergstra, Yamins, and Cox]{bergstra2013making}
J.~Bergstra, D.~Yamins, and D.~D. Cox.
\newblock Making a science of model search: Hyperparameter optimization in
  hundreds of dimensions for vision architectures.
\newblock 2013.

\bibitem[{Bordes} et~al.(2014){Bordes}, {Chopra}, and {Weston}]{qa}
A.~{Bordes}, S.~{Chopra}, and J.~{Weston}.
\newblock Question answering with subgraph embeddings.
\newblock \emph{arXiv e-prints}, 2014.

\bibitem[Cortes and Vapnik(1995)]{cortes1995support}
C.~Cortes and V.~Vapnik.
\newblock Support-vector networks.
\newblock \emph{Machine learning}, 20\penalty0 (3):\penalty0 273--297, 1995.

\bibitem[Cover and Hart(1967)]{cover1967nearest}
T.~Cover and P.~Hart.
\newblock Nearest neighbor pattern classification.
\newblock \emph{IEEE transactions on information theory}, 13\penalty0
  (1):\penalty0 21--27, 1967.

\bibitem[{Deng} et~al.(2009){Deng}, {Dong}, {Socher}, {Li}, {Li}, and
  {Li}]{imagenet}
J.~{Deng}, W.~{Dong}, R.~{Socher}, L.~{Li}, K.~{Li}, and F.~{Li}.
\newblock Imagenet: A large-scale hierarchical image database.
\newblock In \emph{2009 IEEE Conference on Computer Vision and Pattern
  Recognition}, pages 248--255, 2009.

\bibitem[Domhan et~al.(2015)Domhan, Springenberg, and
  Hutter]{domhan2015speeding}
T.~Domhan, J.~T. Springenberg, and F.~Hutter.
\newblock Speeding up automatic hyperparameter optimization of deep neural
  networks by extrapolation of learning curves.
\newblock In \emph{Twenty-Fourth International Joint Conference on Artificial
  Intelligence}, 2015.

\bibitem[Elsken et~al.(2018)Elsken, Metzen, and Hutter]{elsken2018neural}
T.~Elsken, J.~H. Metzen, and F.~Hutter.
\newblock Neural architecture search: A survey.
\newblock \emph{arXiv preprint arXiv:1808.05377}, 2018.

\bibitem[Farabet et~al.(2013)Farabet, Couprie, Najman, and Lecun]{image_2}
C.~Farabet, C.~Couprie, L.~Najman, and Y.~Lecun.
\newblock Learning hierarchical features for scene labeling.
\newblock In \emph{IEEE Trans Pattern Anal Mach Intell}. IEEE, 2013.

\bibitem[Fisher(1936)]{fisher1936use}
R.~A. Fisher.
\newblock The use of multiple measurements in taxonomic problems.
\newblock \emph{Annals of eugenics}, 7\penalty0 (2):\penalty0 179--188, 1936.

\bibitem[{Hinton} et~al.(2012){Hinton}, {Deng}, {Yu}, {Dahl}, {Mohamed},
  {Jaitly}, {Senior}, {Vanhoucke}, {Nguyen}, {Sainath}, and
  {Kingsbury}]{speech_1}
G.~{Hinton}, L.~{Deng}, D.~{Yu}, G.~E. {Dahl}, A.~{Mohamed}, N.~{Jaitly},
  A.~{Senior}, V.~{Vanhoucke}, P.~{Nguyen}, T.~N. {Sainath}, and
  B.~{Kingsbury}.
\newblock Deep neural networks for acoustic modeling in speech recognition: The
  shared views of four research groups.
\newblock \emph{IEEE Signal Processing Magazine}, 2012.

\bibitem[Hutter et~al.(2014)Hutter, Hoos, and Leyton-Brown]{fanova}
F.~Hutter, H.~Hoos, and K.~Leyton-Brown.
\newblock An efficient approach for assessing hyperparameter importance.
\newblock In \emph{Proceedings of the 31st International Conference on Machine
  Learning}, pages 754--762. PMLR, 2014.

\bibitem[{Iandola} et~al.(2016){Iandola}, {Han}, {Moskewicz}, {Ashraf},
  {Dally}, and {Keutzer}]{squeezenet}
F.~N. {Iandola}, S.~{Han}, M.~W. {Moskewicz}, K.~{Ashraf}, W.~J. {Dally}, and
  K.~{Keutzer}.
\newblock {SqueezeNet: AlexNet-level accuracy with 50x fewer parameters and
  less than 0.5MB model size}.
\newblock \emph{arXiv e-prints}, 2016.

\bibitem[{Jasper} et~al.(2012){Jasper}, {Hugo}, and {Ryan}]{prac-bayesian}
S.~{Jasper}, L.~{Hugo}, and P.~A. {Ryan}.
\newblock Practical bayesian optimization of machine learning algorithms.
\newblock \emph{Bartlett et al. [8], pp. 2960–2968}, 2012.

\bibitem[Kerschke et~al.(2019)Kerschke, Hoos, Neumann, and
  Trautmann]{kerschke2019automated}
P.~Kerschke, H.~H. Hoos, F.~Neumann, and H.~Trautmann.
\newblock Automated algorithm selection: Survey and perspectives.
\newblock \emph{Evolutionary computation}, 27\penalty0 (1):\penalty0 3--45,
  2019.

\bibitem[Kohavi and John(1995)]{diff_datasets}
R.~Kohavi and G.~H. John.
\newblock Automatic parameter selection by minimizing estimated error.
\newblock In \emph{Proceedings of the Twelfth International Conference on
  International Conference on Machine Learning}, pages 304--312. Morgan
  Kaufmann Publishers Inc., 1995.

\bibitem[Kotthoff et~al.(2017)Kotthoff, Thornton, Hoos, Hutter, and
  Leyton-Brown]{kotthoff2017autoWEKA}
L.~Kotthoff, C.~Thornton, H.~H. Hoos, F.~Hutter, and K.~Leyton-Brown.
\newblock Auto-weka 2.0: Automatic model selection and hyperparameter
  optimization in weka.
\newblock \emph{The Journal of Machine Learning Research}, 18\penalty0
  (1):\penalty0 826--830, 2017.

\bibitem[Koutsoukas et~al.(2017)Koutsoukas, Monaghan, Li, and
  Huan]{koutsoukas2017deep}
A.~Koutsoukas, K.~J. Monaghan, X.~Li, and J.~Huan.
\newblock Deep-learning: investigating deep neural networks hyper-parameters
  and comparison of performance to shallow methods for modeling bioactivity
  data.
\newblock \emph{Journal of cheminformatics}, 9\penalty0 (1):\penalty0 42, 2017.

\bibitem[Krizhevsky et~al.(2012)Krizhevsky, Sutskever, and Hinton]{image_1}
A.~Krizhevsky, I.~Sutskever, and G.~E. Hinton.
\newblock Imagenet classification with deep convolutional neural networks.
\newblock In \emph{Advances in Neural Information Processing Systems 25}, pages
  1097--1105. Curran Associates, Inc., 2012.

\bibitem[LeCun et~al.(2015)LeCun, Bengio, and Hinton]{deeplearning}
Y.~LeCun, Y.~Bengio, and G.~Hinton.
\newblock Deep learning.
\newblock \emph{Nature}, 2015.

\bibitem[Leung et~al.(2014)Leung, Xiong, Lee, and Frey]{bioinformatics}
M.~K.~K. Leung, H.~Y. Xiong, L.~J. Lee, and B.~J. Frey.
\newblock Deep learning of the tissue-regulated splicing code.
\newblock \emph{Bioinformatics (Oxford, England)}, 2014.

\bibitem[Liu et~al.(2018)Liu, Simonyan, and Yang]{liu2018darts}
H.~Liu, K.~Simonyan, and Y.~Yang.
\newblock Darts: Differentiable architecture search.
\newblock \emph{arXiv preprint arXiv:1806.09055}, 2018.

\bibitem[{Lucic} et~al.(2017){Lucic}, {Kurach}, {Michalski}, {Gelly}, and
  {Bousquet}]{gans}
M.~{Lucic}, K.~{Kurach}, M.~{Michalski}, S.~{Gelly}, and O.~{Bousquet}.
\newblock {Are GANs Created Equal? A Large-Scale Study}.
\newblock \emph{arXiv e-prints}, 2017.

\bibitem[Ma et~al.(2015)Ma, Sheridan, Liaw, Dahl, and Svetnik]{drug}
J.~Ma, R.~P. Sheridan, A.~Liaw, G.~E. Dahl, and V.~Svetnik.
\newblock Deep neural nets as a method for quantitative structure-activity
  relationships.
\newblock \emph{Journal of Chemical Information and Modeling}, 2015.

\bibitem[{Melis} et~al.(2017){Melis}, {Dyer}, and {Blunsom}]{nlp-reproduce}
G.~{Melis}, C.~{Dyer}, and P.~{Blunsom}.
\newblock {On the State of the Art of Evaluation in Neural Language Models}.
\newblock \emph{arXiv e-prints}, 2017.

\bibitem[Olkin(1960)]{levene}
I.~Olkin.
\newblock {Contributions to probability and statistics; essays in honor of
  Harold Hotelling}.
\newblock \emph{Stanford, Calif., Stanford University Press, 1960.}, 1960.

\bibitem[Pham et~al.(2018)Pham, Guan, Zoph, Le, and Dean]{pham2018efficient}
H.~Pham, M.~Y. Guan, B.~Zoph, Q.~V. Le, and J.~Dean.
\newblock Efficient neural architecture search via parameter sharing.
\newblock \emph{arXiv preprint arXiv:1802.03268}, 2018.

\bibitem[Reimers and Gurevych(2017)]{reimers2017optimal}
N.~Reimers and I.~Gurevych.
\newblock Optimal hyperparameters for deep lstm-networks for sequence labeling
  tasks.
\newblock \emph{arXiv preprint arXiv:1707.06799}, 2017.

\bibitem[Rish(2001)]{rish2001empirical}
I.~Rish.
\newblock An empirical study of the naive bayes classifier.
\newblock In \emph{IJCAI 2001 workshop on empirical methods in artificial
  intelligence}, volume~3, pages 41--46, 2001.

\bibitem[{Sainath} et~al.(2013){Sainath}, {Kingsbury}, {Mohamed}, {Dahl},
  {Saon}, {Soltau}, {Beran}, {Aravkin}, and {Ramabhadran}]{speech_2}
T.~N. {Sainath}, B.~{Kingsbury}, A.~{Mohamed}, G.~E. {Dahl}, G.~{Saon},
  H.~{Soltau}, T.~{Beran}, A.~Y. {Aravkin}, and B.~{Ramabhadran}.
\newblock Improvements to deep convolutional neural networks for lvcsr.
\newblock \emph{arXiv e-prints}, 2013.

\bibitem[Shahriari et~al.(2016)Shahriari, Swersky, Wang, Adams, and
  de~Freitas]{taking-humanout-of-loop}
B.~Shahriari, K.~Swersky, Z.~Wang, R.~P. Adams, and N.~de~Freitas.
\newblock Taking the human out of the loop: A review of bayesian optimization.
\newblock \emph{Proceedings of the IEEE}, pages 148--175, 2016.

\bibitem[Smith(2018)]{smith2018disciplined}
L.~N. Smith.
\newblock A disciplined approach to neural network hyper-parameters: Part
  1--learning rate, batch size, momentum, and weight decay.
\newblock \emph{arXiv preprint arXiv:1803.09820}, 2018.

\bibitem[Spearman(2008)]{spearman}
C.~Spearman.
\newblock \emph{Spearman Rank Correlation Coefficient}, pages 502--505.
\newblock Springer New York, 2008.

\bibitem[van Rijn and Hutter(2018)]{across_datasets}
J.~N. van Rijn and F.~Hutter.
\newblock Hyperparameter importance across datasets.
\newblock In \emph{Proceedings of the 24th ACM SIGKDD International Conference
  on Knowledge Discovery and Data Mining}, pages 2367--2376, 2018.

\bibitem[Zhou et~al.(2018)Zhou, Cahya, Combs, Nicolaou, Wang, Desai, and
  Shen]{zhou2018exploring}
Y.~Zhou, S.~Cahya, S.~A. Combs, C.~A. Nicolaou, J.~Wang, P.~V. Desai, and
  J.~Shen.
\newblock Exploring tunable hyperparameters for deep neural networks with
  industrial adme data sets.
\newblock \emph{Journal of chemical information and modeling}, 59\penalty0
  (3):\penalty0 1005--1016, 2018.

\bibitem[Zoph and Le(2016)]{zoph2016neural}
B.~Zoph and Q.~V. Le.
\newblock Neural architecture search with reinforcement learning.
\newblock \emph{arXiv preprint arXiv:1611.01578}, 2016.

\end{thebibliography}
\end{document}


\maketitle

\section{Hyperparameters used}
\label{sec:hp_explanation}
The hyperparameters involved in the experiment are divided into two categories, mandatory and non-mandatory. The mandatory hyperparameters are necessary and required to train the model where as non-mandatory hyperparameters are subjected to the choice of your optimizer algorithm. If not specified, the default values for these non-mandatory hyperparameters are used to train the model. However, dividing these hyperparameters into these categories does not necessarily mean that mandatory hyperparameters are more important and impact the performance of the model more than other category.
A brief information and default-values (used by PyTorch) is provided below for each Hyperparameter:
\subsection{Mandatory hyperparameters}
    \begin{enumerate}
        \item Epoch: 
            epoch is defined as the number of iterations over the dataset. Minimum value is 1.
        \item Batch size: 
            batch size is the total number of training samples present in a single batch. Minimum value is 1.
        \item Loss function: 
            Every model learn by means of a loss function. It is a method of evaluating how well the model has learnt the given data. If the prediction deviates too much from truth results, loss function is usually high. There are different types of loss function available to choose from and each options is explained below:
            \begin{enumerate}
                \item Cross entropy:
                    Cross entropy loss, or log loss, measures the performance of a classification model whose output is a probability value between 0 and 1. Cross-entropy loss increases as the predicted probability diverges from the actual label. So predicting a probability of 0.1 when the actual observation label is 1 would be bad and result in a high loss value. A perfect model would have a log loss of 0.
                \item L1 loss:
                    It is also known as L1-norm loss function. L1 loss is basically minimizing the sum of the absolute differences between the target value and the estimated values. A perfect model would have a log loss of 0.
                \item Mean squared loss:
                    It is also known as L2-norm loss function. It is basically minimizing the sum of the square of the differences between the target value and the estimated values.
                \item Negative likelihood:
                    The negative log-likelihood becomes high at smaller values, where it can reach infinite loss, and becomes less at larger values.
            \end{enumerate}
        
        \item Optimizer  algorithm:
            During the training process, we tweak and change the parameters (weights) of our model to try and minimize that loss function, and make our predictions as correct as possible. But how exactly do you do that? How do you change the parameters of your model, by how much, and when? Optimizer together with the loss function and model parameters by updating the model in response to the output of the loss function. In simpler terms, optimizer's shape and mold your model into its most accurate possible form. The available choices for the Optimizer are explained below:
            \begin{enumerate}
                \item Adam optimizer
                \item Adadelta
                \item Averaged stochastic gradient
                \item RMSprop
                \item Stochastic gradient
                \item Adagrad
            \end{enumerate}
    \end{enumerate}

\subsection{Non-Mandatory hyperparameters}
    \begin{enumerate}
        \item Learning rate: 
            Learning rate is a hyper-parameter that controls how much we are adjusting the weights of our network with respect the loss gradient. It is used by all optimization algorithms mentioned above. (Default value: 0.001)
        \item Weight decay: 
            When training neural networks, it is common to use weight decay, where after each update, the weights are multiplied by a factor slightly less than 1. This prevents the weights from growing too large, and can be seen as gradient descent on a quadratic regularization term. It is used by all optimization algorithms mentioned above. (Default value: 0)
        \item Rho: 
            Rho is a coefficient used for computing a running average of squared gradients. (Default value: 0.9)
        \item Lambda: 
            Lambda is used as a decay term for gradient update. (Default value: 0)

        \item Alpha: 
            Alpha is used as a smoothing constant.(Default value: 0.99)
         \item Epsilon: 
            Sometimes the value of gradient could be really close to 0. Then, the value of the weights could blow up. To prevent the gradients from blowing up, epsilon could be included in the denominator. (Default value: 0.00001)
            
        \item Momentum:
            Momentum helps accelerate gradients vectors in the right directions, thus leading to faster converging. (Default value: 0)
        \item Learning rate decay:  
            It specifies the rate of decay for your learning rate. (Default value: 0)
        \item Initial accumulator value: It defines the starting values for accumulate gradient and accumulate updates. (Default value: 0)
        \item Beta1 and Beta2: 
            These are the coefficients used for computing running averages of gradient and its square. (Default value: 0.9, 0.999)
   \end{enumerate}

\section{User study screenshots}

\begin{figure}[htpb]
\centering
 \captionsetup{width=\columnwidth}

\includegraphics[width=\columnwidth]{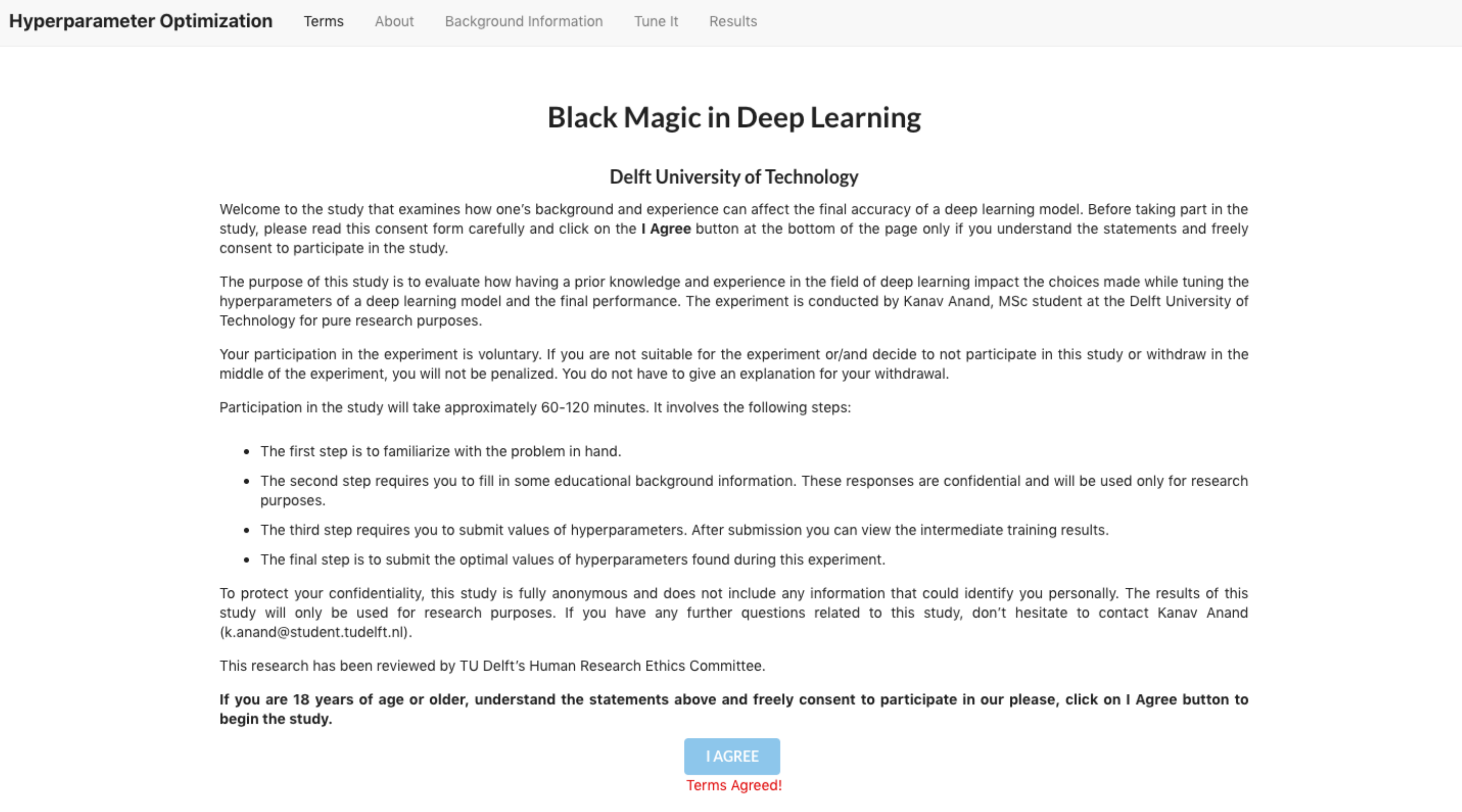}
\caption{The terms and condition webpage used for the user study. It clearly states all expectations and requirements for participants.}
\label{fig:terms}    
\end{figure}
\newpage

\begin{figure}[htpb]
\centering
 \captionsetup{width=\columnwidth}

\includegraphics[width=\columnwidth]{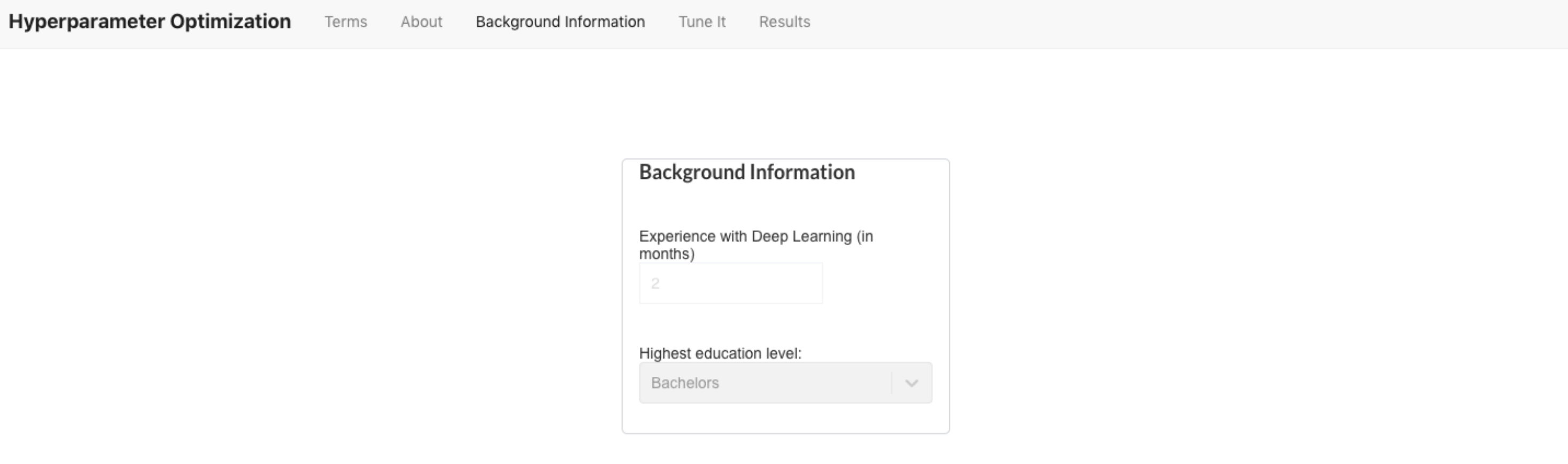}
\caption{The background information page is shown above. It asks the participants to enter their relevant experience in the field of deep learning and their highest completed education level. Once the information is submitted by the user, anonymous user id is generated to identify every unique participant.}
\label{fig:bg}  
\end{figure}
\newpage

\begin{figure}[htpb]
\centering
 \captionsetup{width=\columnwidth}

\includegraphics[width=\columnwidth]{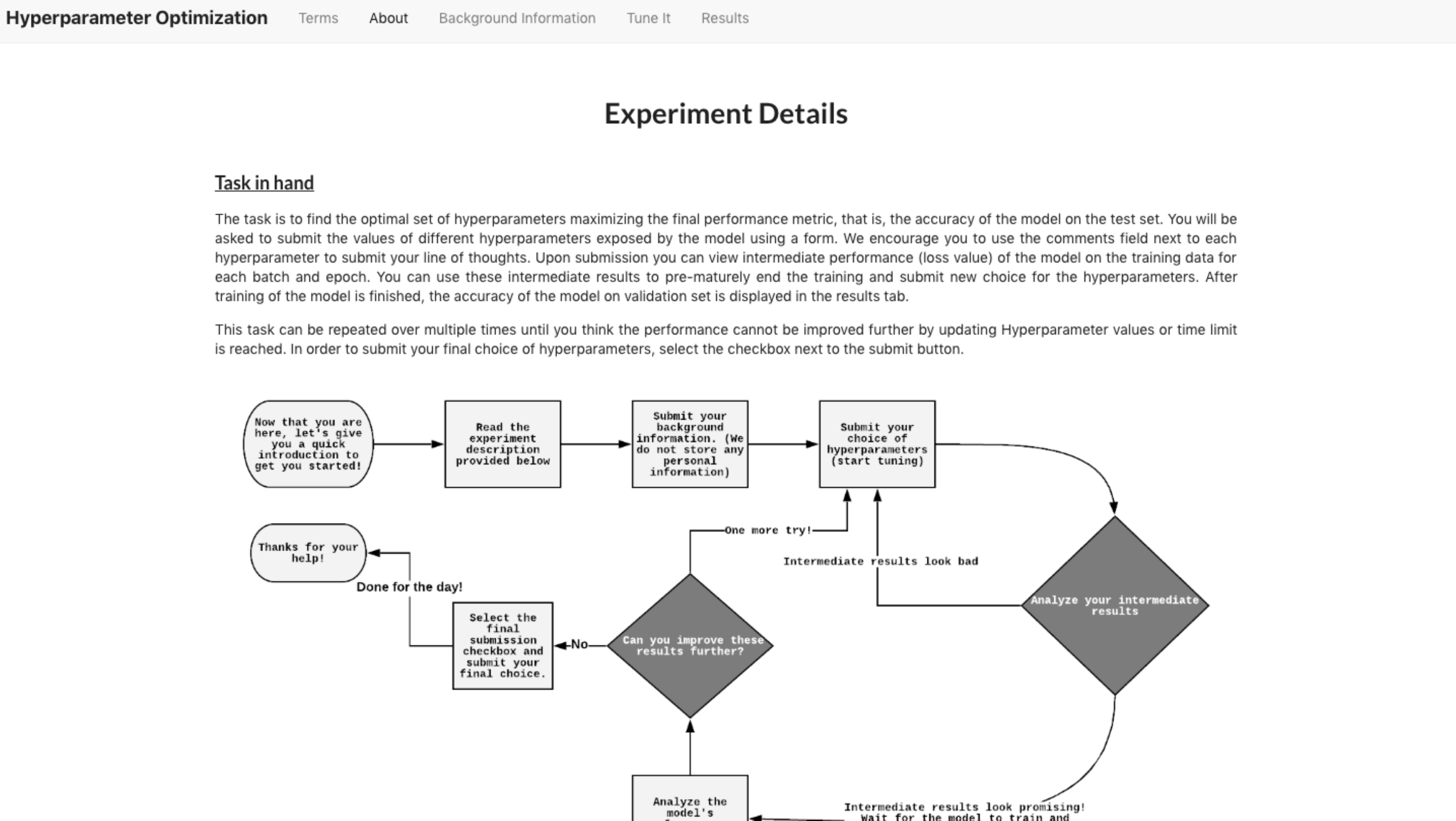}
\caption{The experiment details page provides the detailed information of the user study. It starts with the flow diagram of the user study and follows it up with the explanation of every hyperparameter used for the experiment.}
\label{fig:ed}  
\end{figure}

\newpage
\begin{figure}[htpb]
\centering
 \captionsetup{width=\columnwidth}

\includegraphics[width=\columnwidth]{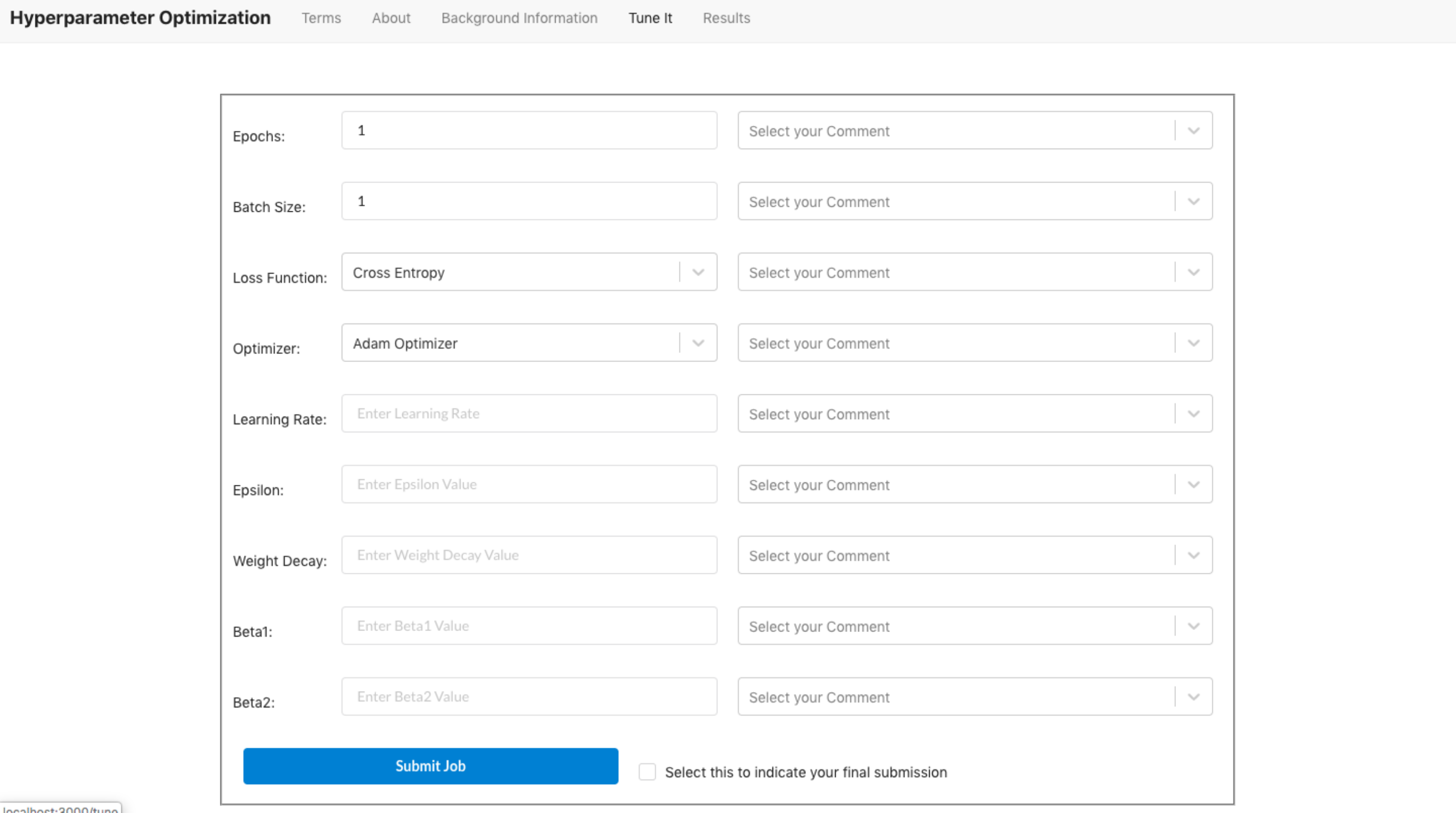}
\caption{The hyperparameter submission form is used by the participant to submit the values for hyperparameters and their corresponding comments. Once a hyperparameter configuration is submitted, the user can view intermediate results and early stop the training of the model.}
\label{fig:hp}  
\end{figure}
